\pdfoutput=1

\documentclass[11pt]{article}

\usepackage{ACL2023}

\usepackage{microtype}

\usepackage{inconsolata}

\usepackage[utf8]{inputenc} 
\usepackage[T1]{fontenc}    
\usepackage{hyperref}       
\usepackage{url}            
\usepackage{booktabs}       
\usepackage{amsfonts}       
\usepackage{nicefrac}       
\usepackage{microtype}      
\usepackage{xcolor}         

\usepackage{times}
\usepackage{latexsym}

\usepackage{graphicx}
\usepackage{subfigure}
\usepackage{booktabs} 
\usepackage{lipsum}
\usepackage{amsmath}
\usepackage{amssymb, amsthm, latexsym}
\usepackage{multirow}

\usepackage{nicefrac}

\usepackage[frozencache,cachedir=.]{minted}

\newminted{python}{%
    fontsize=\fontsize{8.25pt}{8.25pt}\selectfont,
}

\title{\textsc{Petals}: Collaborative Inference and Fine-tuning of Large Models}

\author{
  Alexander Borzunov\thanks{\ \ Equal contribution. Correspondence to:\newline \texttt{borzunov.alexander@gmail.com}}\\
  HSE University, Yandex \\\And
  Dmitry Baranchuk$^*$ \\
  Yandex \\\And
  Tim Dettmers$^*$ \\
  University of Washington\\\AND
  Max Ryabinin$^*$ \\
  HSE University, Yandex \\\And
  Younes Belkada$^*$ \\
  Hugging Face, ENS Paris-Saclay \\\And
  Artem Chumachenko \\
  Yandex \\\AND
  Pavel Samygin \\
  Yandex School of Data Analysis \\\And
  Colin Raffel \\
  Hugging Face
}

\begin{document}
\maketitle
\begin{abstract}
Many NLP tasks benefit from using large language models (LLMs) that often have more than 100 billion parameters. With the release of BLOOM-176B and OPT-175B, everyone can download pretrained models of this scale. Still, using these models requires high-end hardware unavailable to many researchers. In some cases, LLMs can be used more affordably via RAM offloading or hosted APIs. However, these techniques have innate limitations: offloading is too slow for interactive inference, while APIs are not flexible enough for research that requires access to weights, attention or logits. In this work, we propose \textsc{Petals}\footnote{\textsc{Petals} source code and documentation are available at \texttt{\href{https://petals.ml}{https://petals.ml}}} --- a~system for inference and fine-tuning of large models collaboratively by joining the resources of multiple parties. We demonstrate that this strategy outperforms offloading for very large models, running inference of BLOOM-176B on consumer GPUs with $\approx$~1 step per second, which is enough for many interactive LLM applications. Unlike most inference APIs, \textsc{Petals} also natively exposes hidden states of served models, allowing to train and share custom model extensions based on efficient fine-tuning methods.
\end{abstract}

\section{Introduction}


In recent years, the NLP community has found that pretrained language models can solve many practical tasks, through either fine-tuning~\citep{gpt} or simple prompting~\citep{gpt3}. Furthermore, performance tends to improve as scale increases~\citep{gpt2, kaplan2020scaling}. Following this trend, modern language models often have hundreds of billions of parameters~\citep{gpt3,gopher,pangua,hyperclova}\nocite{switch,jurrasic,Lepikhin2020GShardSG,glam}. Several research groups released pretrained LLMs with over 100B parameters~\citep{opt,yalm,zeng2020glm}\nocite{gpt,gpt-neox-20b}. Most recently, the BigScience project has released BLOOM, a 176 billion parameter model supporting 46 natural and 13 programming languages~\citep{bloom}.


While the public availability of 100B+ parameter models makes them easier to access, they remain difficult to use for the majority of researchers and practitioners due to memory and computational costs. For instance, OPT-175B and BLOOM-176B need over 350 GB accelerator memory for inference and significantly more for fine-tuning. As a result, these LLMs usually require multiple high-end GPUs or multi-node clusters\nocite{megatron2} to be run. Both of these options are extremely expensive, which limits research and potential applications of LLMs.


Several recent works aim to democratize LLMs
by ``offloading'' model parameters to slower but cheaper memory (RAM or SSD), then running them on the accelerator layer by layer~\citep{l2l,zerooffload}\nocite{accelerate}.
This method allows running LLMs with a single low-end accelerator by loading parameters from RAM justin-time for each forward pass.
Offloading can be efficient for processing many tokens in parallel, but it has inherently high latency: for example, generating one token at a time with BLOOM-176B takes at least \textit{5.5~seconds} for the fastest RAM offloading setup and 22~seconds for the fastest SSD offloading. In addition, many computers do not have enough RAM to offload 175B parameters.

\begin{figure*}[t]
    \centering
    \vspace{-16pt}
    \includegraphics[width=\linewidth]{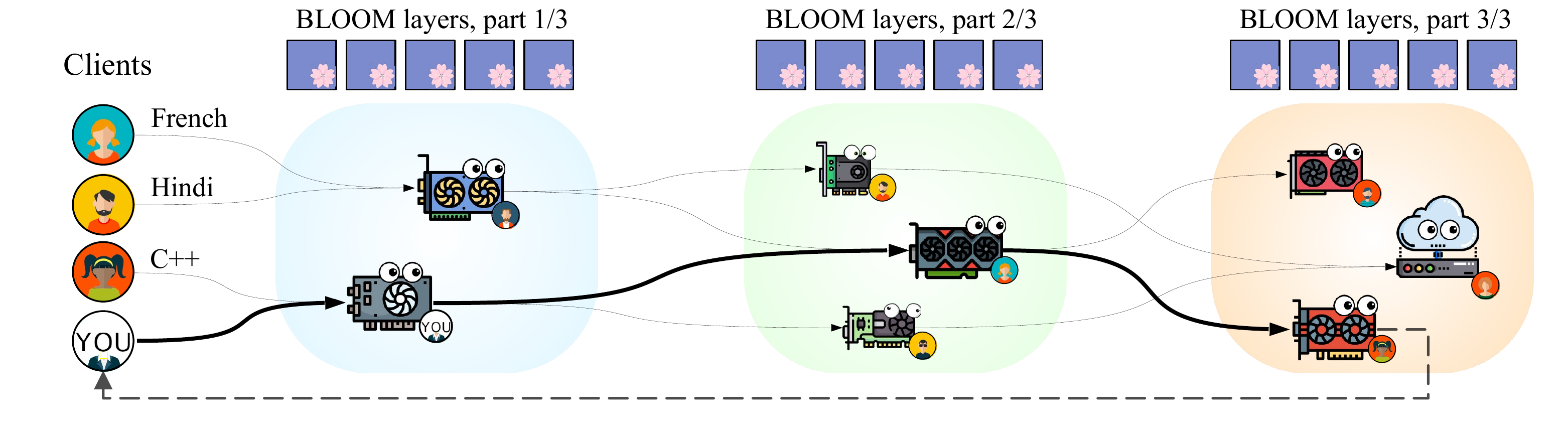}
    \vspace{-16pt}
    \caption{An overview of \textsc{Petals}. Some participants (\textit{clients}) want to use a pretrained language model to solve various tasks involving processing texts in natural (e.g., French, Hindi) or programming (e.g., C++) languages. They do it with help of other participants (\textit{servers}), who hold various subsets of model layers on their GPUs. Each client chooses a sequence of servers so that it performs an inference or fine-tuning step in the least amount of time.}
    \label{fig:overview}
    \vspace{-12pt}
\end{figure*}

Another way to make LLMs more accessible is through public inference APIs, where one party hosts the model and lets others query it over the Internet~\citep{openai-api,jurrasic,forefront}. Since most of the engineering work is done by the API owner, this is a relatively user-friendly option.
However, APIs are often not flexible enough for research use: there is no way to change the model control flow or access internal states. On top of that, current API pricing can make some research projects prohibitively expensive~\citep{tfew}.

In this work, we explore an alternative strategy inspired by crowdsourced distributed training of neural networks from scratch~\citep{hivemind_dmoe}. We introduce \textsc{Petals}, a platform that allows multiple users to collaborate and perform inference and fine-tuning of large language models over the Internet.
Each participant runs a server, a client or both. A server hosts a subset of model layers (typically, Transformer blocks) and handles requests from clients.
A client can form a chain of pipeline-parallel consecutive servers to run the inference of the entire model (Section~\ref{sect:design_inference}).
Aside from inference, participants can fine-tune the model through parameter-efficient training methods like adapters \citep{houlsby2019parameter} or prompt tuning \citep{ptune-lester} or by training entire layers (Section~\ref{sect:design_training}). Once trained, submodules can be shared on a model hub (Section~\ref{sect:design_ecosystem}), where others can use them for inference or further training.
We demonstrate that existing 100B+ models can run efficiently in this setting with the help of several optimizations: dynamic quantization, prioritizing low-latency connections, and load balancing between servers (Section~\ref{sect:internals}).



\section{Design and use cases}

Practical usage of large language models can be broadly divided into two main scenarios: inference and parameter-efficient adaptation to downstream tasks. In this section, we outline the design of \textsc{Petals}, showing how it handles both scenarios and also allows easily sharing trained adapters between the users of the system.

\subsection{Inference of billion-scale models}\label{sect:design_inference}

When generating tokens, a client stores the model's token embeddings (which typically comprise a small fraction of the total parameter count and can fit in RAM in most modern laptops, servers, and workstations) locally and relies on servers to run Transformer blocks. Each server holds several \textit{consecutive} blocks, the number of which depends on the server's available GPU memory.
Before each inference session, the client finds a chain of servers that collectively hold all model layers.

Once the chain is formed, the client uses the local embedding layer to look up embedding vectors for prefix tokens, then sends those vectors to servers and receives new representations. Once the client obtains the outputs of the final block, it computes next token probabilities and repeats this process.

While the session is active, servers store attention keys and values from past client inputs and use them for subsequent inference steps. Clients also store past inputs to each server so that if any server fails or goes offline, another one can quickly take its place. The procedure for finding servers and recovering from failures is detailed in Section~\ref{sect:networking}.

\paragraph{Client-side API.} To generate tokens with \textsc{Petals}, one first creates an \textit{inference session}. An inference session iteratively takes inputs as PyTorch tensors, runs them through all Transformer blocks and returns final representations as PyTorch tensors. Under the hood, sessions form server chains, hold cache, and recover from server failures in a way that is transparent to the user. An example of using an inference session is shown in Figure~\ref{fig:infernce_snippet}.


\begin{figure}[tb]
\small
\begin{pythoncode}
# Initialize distributed BLOOM model
model = DistributedBloomForCausalLM \
    .from_pretrained("bigscience/bloom-petals")
input_ids = tokenizer(prefix_text)

with model.inference_session() as session:
    # Session maintains a set of servers that
    # store attention KV from previous steps
    for _ in range(sequence_length):
        # Compute the word embeddings locally
        hid = model.word_embeddings(input_ids)
        # Run distributed Transformer blocks,
        # store attention KV for future steps
        hid = session.step(hid)
        # Sample the next token locally
        probs = model.lm_head(hid)
        input_ids = sample_next_token(probs)
\end{pythoncode}
    \vspace{-5pt}
    \caption{A basic PyTorch code snippet for generation with a distributed BLOOM-176B model.}
    \label{fig:infernce_snippet}
    \vspace{-10pt}
\end{figure}

\paragraph{System requirements.} For BLOOM-176B inference, clients need at least 12 GB RAM, most of which is used to store 3.6B embedding parameters. We recommend at least 25 Mbit/s bidirectional bandwidth to avoid bottlenecks in network transfers. Simple greedy inference can use any CPU that runs PyTorch, but more advanced algorithms (e.g., beam search) may require a GPU.

In turn, servers need at least 16 GB of CPU RAM, 100 Mbit/s bandwidth and a GPU with at least 8~GB of memory.

\begin{figure}[tb]
    \centering
    \includegraphics[width=0.9\linewidth]{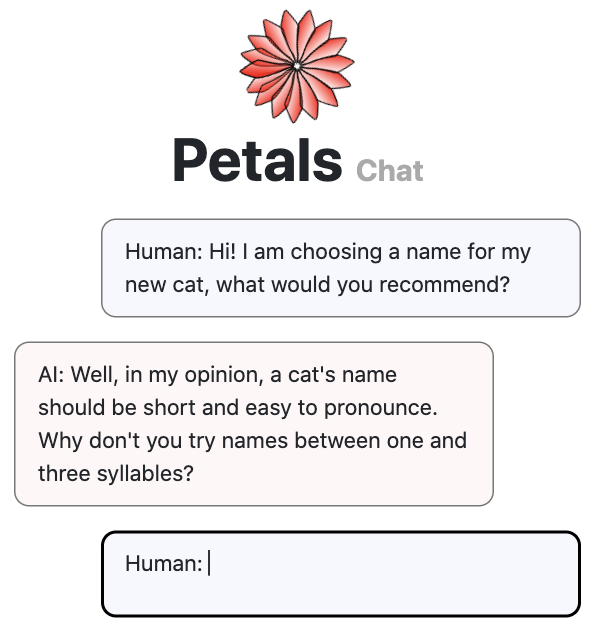}
    \caption{A chat application that runs BLOOM-176B or BLOOMZ-176B over the \textsc{Petals} swarm, available at \texttt{\href{https://chat.petals.ml}{https://chat.petals.ml}}}
    \label{fig:chat_app}
    \vspace{-10pt}
\end{figure}

\paragraph{Chat application.} We also provide an example application that lets users chat with LLMs in a messenger-like user interface (see Figure~\ref{fig:chat_app}). The application supports BLOOM-176B and BLOOMZ-176B, a version of BLOOM fine-tuned to better perform in the zero-shot regime~\cite{bloomz}. The application is comprised of the \textit{frontend} and the \textit{backend}.
The frontend is a web page that allows users to communicate with the model by prompting it with text and receiving the generated output.
The backend is a Flask web server that uses the \textsc{Petals} client to run inference over the swarm. It accepts requests via HTTP or Websocket protocols, so anyone can develop their own applications using our backend for inference.


\subsection{Training for downstream tasks}\label{sect:design_training}

While LLMs achieve high quality on many problems with simple prompt engineering~\citep{gpt3}, they often need training to achieve the best results. Traditionally, this is done by fine-tuning all model parameters on the downstream task.
However, for very large models, this strategy becomes impractical due to hardware requirements. For example, fine-tuning BLOOM-176B with Adam would require almost 3~TB of GPU memory to store model, gradients, and optimizer states.

To combat this issue, the NLP community has developed \textit{parameter-efficient fine-tuning} methods that keep most of the pretrained model intact. Some of them~\citep{sung2021training,guo2021parameter} choose a subset of existing parameters, others~\citep{hu2021lora, houlsby2019parameter, ptune-liu, ptune-lester, ptune-v2, tfew} augment the model with extra trainable weights.

Despite their lower memory requirements, parameter-efficient approaches are often competitive with full model fine-tuning \citep{hu2021lora,ptune-v2,yong_adapting} and even outperform it in low-data regimes~\citep{2205.05638}. Another appealing property of these approaches for our use-case is that they allow rapidly switching a pretrained LLM between different uses.


\paragraph{Distributed fine-tuning.} The core principle of fine-tuning in a distributed network is that clients ``own'' trained parameters while servers host original pretrained layers. Servers can run backpropagation through their layers and return gradients with respect to activations, but they \textit{do not update the server-side parameters}. Thus, clients can simultaneously run different training tasks on the same set of servers without interfering with one another.



To illustrate this principle, we first review an example of soft prompt-tuning for text classification and then generalize it to other methods and tasks.
Similarly to Section~\ref{sect:design_inference}, clients store the embedding layers locally and rely on servers to compute the activations of Transformer blocks. In this fine-tuning scenario, a client needs to store trainable soft prompts (task-specific input embeddings) and a linear classification head. 

For each training batch, the client routes its data through a chain of remote servers to compute sentence representations, then obtains predictions with the classifier head and computes the cross-entropy loss.
During backpropagation, the client runs its data through the same chain of servers in reverse order to compute gradients for the learned prompt vectors. Having obtained those gradients, the client can use a regular PyTorch optimizer to update the parameters of both the head and the prompts, then proceed to the next minibatch.

\paragraph{User interface.} To allow users greater flexibility in their training workloads, we made distributed backpropagation module compatible with the PyTorch Autograd engine. Like in the inference stage, this module handles fault tolerance and load balancing transparently to the user while allowing them to access intermediate activations and insert custom PyTorch modules. Figure~\ref{fig:training_snippet} shows an example training code snippet.

This interface can also support other popular parameter-efficient fine-tuning algorithms, such as LoRA~\citep{hu2021lora} or prefix tuning~\citep{li-liang-2021-prefix}. Finally, users can insert custom local modules after some of the  existing blocks, which could allow use-cases like retrieval-augmented generation~\citep{retro,rag}.

\begin{figure}[tb]
\small
\begin{pythoncode}
# Use distributed BLOOM with soft prompts
model = AutoModelForSequenceClassification \
    .from_pretrained(
        "bigscience/bloom-petals",
        tuning_mode="ptune", pre_seq_len=5)
# Define optimizer for prompts and linear head
opt = torch.optim.AdamW(model.parameters())

for input_ids, labels in data_loader:
    # Forward pass with local & remote layers
    out = model.forward(input_ids)
    loss = cross_entropy(out.logits, labels)

    # Distributed backward w.r.t. local params
    loss.backward() # Compute prompts.grad
    opt.step() # Update local params only
    opt.zero_grad()
\end{pythoncode}
\vspace{-5pt}
\caption{A basic PyTorch code of soft prompt tuning for sequence classification with \textsc{Petals}.}
\label{fig:training_snippet}
\vspace{-10pt}
\end{figure}

\subsection{Sharing and reusing trained modules}
\label{sect:design_ecosystem}

Although most fine-tuned extensions for pretrained models can be easily shared as-is, simplifying the workflow for sharing these extensions enables users to more easily adapt the model to their target scenario. Indeed, existing model hubs~\citep{wolf-etal-2020-transformers, tfhub, torchhub} have gained immense popularity due to many supported models and ease of use, especially when vetting different pretrained models for a given problem. One particularly relevant project is AdapterHub~\citep{adapterhub}, a repository of trained adapters accompanied by a library with implementations of different adaptation methods. While \textsc{Petals} does not depend on AdapterHub, it is possible to leverage this library for training adapters in the distributed setting.
Instead, we support sharing modules trained by users via the Hugging Face Hub (also used as a backend by AdapterHub). Its infrastructure and the corresponding open source library simplify the learning process for users already familiar with the ecosystem. Because the primary navigation mechanism on the Hugging Face Hub are tags that have been applied to uploaded modules, a user only needs to the task it was trained on and the model upon which the adapter was built. Uploading the weights and the code of the fine-tuned module is done by committing them to a Git repository.
When navigating the Hub, users can choose the most suitable adapters by filtering the list of all available modules by the required tags.
\section{Internal structure and optimizations}\label{sect:internals}

One of the primary considerations for distributed inference is its performance. It can be broken down into three main aspects: computation speed (5-year-old gaming GPU vs. new data center GPU), communication delay due to distance between nodes (intercontinental vs. local), and communication delay due to bandwidth (10 Mbit/s vs. 10 Gbit/s).

In terms of raw FLOPs, even consumer-grade GPUs like GeForce RTX 3070 could run a complete inference step of BLOOM-176B in less than a second~\citep{ga102-datasheet}. However, the GPU memory can only hold a small fraction of model layers: running na\"ively would require 44 RTX 3070 GPUs and 44 communication rounds. To make this more efficient, we use quantization to store more parameters per GPU, reducing the number of consecutive devices and communication rounds (Section~\ref{sect:inside_gpu}). On top of that, each client prioritizes nearby servers to make communication rounds faster (Section~\ref{sect:networking}).



\subsection{Large model inference on consumer GPUs}\label{sect:inside_gpu}

We assume that each server has at least 16 GB of CPU RAM, 8 GB of GPU memory. From this assumption, one of the primary considerations is to reduce the model memory footprint, so that each device can hold more Transformer blocks.

For example, BLOOM has 176B parameters, which takes 352 GB of GPU memory in 16-bit precision. Thus, in the worst case, the model is distributed among 352 GB / 8 GB (per server) = 44 nodes. We can reduce both frequency and amount of data transfer in two ways.
First, we can achieve this by compressing the hidden states exchanged between nodes. Second, we can compress the weights to 8-bit precision, reducing the number of nodes required to hold all layers. For BLOOM, this changes the number of required nodes from 44 to 22, which reduces latency in half and decreases the probability of a failure.

\paragraph{Compressing communication buffers.} To send less data between subsequent pipeline stages, we use dynamic blockwise quantization \citep{dettmers2022optimizers}. We apply it to the hidden states before pipeline-parallel communication, as done in \citet{ryabinin2021swarm}. Dynamic blockwise quantization halves the bandwidth requirements without any noticeable effect on generation quality.

\paragraph{Compressing model weights.} We use 8-bit mixed matrix decomposition for matrix multiplication to quantize the weights to 8-bit precision and reduce the memory footprint compared to 16-bit weights, as suggested in \citep{dettmers2022llm}. This decomposition separates hidden states and weights into two portions: about 0.1\% of 16-bit outlier and 99.9\% of 8-bit regular values, which roughly halves the memory footprint.



As shown in Table~\ref{tab:quality}, this method has little effect on LLM quality for major benchmarks.
In terms of inference time, Table~\ref{tab:throughput} demonstrates that quantization has about $5\%$ of overhead with batch size 1 (20 tokens), but becomes negligible for larger batches.


\subsection{Collaborating over the Internet}\label{sect:networking}

\begin{table}[tb]
\centering
\caption{Zero-shot accuracy for OPT-175B and BLOOM-176B with 8-bit and 16-bit weights.\nocite{eval-harness}}
\vspace{-5pt}
\resizebox{\linewidth}{!}{%
\begin{tabular}{lccccc}\toprule
\textbf{Model}            & \textbf{Bits} & \textbf{HellaSwag} & \textbf{LAMBADA} & \textbf{WinoGrande} & \textbf{Avg}             \\\midrule
\multirow{2}{*}{OPT-175B} & 16            & 78.5               & 74.7             & 72.6                & 75.3                     \\
                          & 8             & 78.5               & 74.6             & 71.7                & \multicolumn{1}{r}{74.9} \\\midrule
\multirow{2}{*}{BLOOM}    & 16            & 73.0               & 67.2             & 70.1                & 70.1                     \\
                          & 8             & 72.8               & 68.1             & 70.1                & 70.3                    \\\bottomrule
\end{tabular}}
\label{tab:quality}
\end{table}

\begin{table}[tb]
\centering
\caption{Generation throughput (tokens/s) for BLOOM-176B with 8-bit and 16-bit weights on 8$\times$~A100 GPUs.}
\vspace{-5pt}
\label{tbl:memory_footprint}
\resizebox{0.5\linewidth}{!}{%
\begin{tabular}{lccc}\toprule
\multirow{2}{*}{\bf Weights}& \multicolumn{3}{c}{\bf Batch size} \\\cmidrule{2-4} & \bf 1 & \bf 8 &\bf 32  \\\toprule
16-bit       & 4.18 & 31.3  & 100.6  \\
8-bit        & 3.95 & 29.4  & 95.8\\\bottomrule
\end{tabular}}
\label{tab:throughput}
\vspace{-10pt}
\end{table}

Another challenge is to provide \textit{reliable} inference and training despite nodes joining, leaving or failing at any time. To address this, \textsc{Petals} uses the \texttt{hivemind} library~\citep{hivemind} for decentralized training and custom fault-tolerant protocols for servers and clients.


\paragraph{Server load balancing.} First, we ensure that servers are distributed evenly among Transformer blocks. Formally, servers maximize the total model throughput by choosing the blocks with the worst throughput and eliminating potential bottlenecks.

Each server periodically announces its active blocks to a distributed hash table~\citep{kademlia}. When a new server joins, it uses this information to identify an interval of blocks that contains most blocks with the worst throughput. This interval is always contiguous, since splitting it would harm the inference latency. Once the server has selected its layers, it measures its own throughput (both network and compute) and announces it to the distributed hash table. 

Since peers may leave or fail at any time, all nodes periodically check if launching a rebalancing procedure would significantly improve the overall throughput. If it is the case, they switch layers until the throughput becomes near-optimal. In particular, if all peers serving certain blocks suddenly leave the system, this procedure quickly redistributes the remaining resources to close the emerged gaps.



\paragraph{Client-side routing.} Next, we want clients to be able to find a sequence of servers that run the model in the least amount of time. During generation, clients process one or few tokens at a time; in practice, the inference time is mostly sensitive to the network latency. Thus, clients have to ping nearby servers to measure latency and then find the path with minimal time via beam search. Conversely, during fine-tuning one needs to process a batch of examples in parallel. 
Here, clients can split their batches between multiple servers using the algorithm from~\citet{ryabinin2021swarm}.
If a server fails during training or inference, a client removes it from consideration and reruns routing to find a replacement. During inference, the client sends all previous inputs to the replacement server, so that it has the same attention keys and values.





\begin{table}[t]
\centering
 \caption{Performance of sequential inference steps and parallel forward passes. RTT is the round-trip latency.}
 \vspace{-5px}
\label{tbl:experiments}
\resizebox{\linewidth}{!}{
\setlength{\tabcolsep}{6pt}
\begin{tabular}{ccccc}\toprule
\multirow{3}{*}{\bf{Network}} & \multicolumn{2}{c}{\bf{Single-batch}} & \multicolumn{2}{c}{\bf{Parallel}}\\
& \multicolumn{2}{c}{\bf{inference (steps/s)}} & \multicolumn{2}{c}{\bf{forward (tokens/s)}}\\
\cmidrule{2-5}
& \multicolumn{2}{c}{\bf{Sequence length}} & \multicolumn{2}{c}{\bf{Batch size}}\\
\midrule
\bf{Bandwidth, RTT} & 128 & 2048 & 1 & 64 \\
\midrule
\multicolumn{5}{c}{\textsc{Petals} on 3 physical servers, with one A100 each}\\
\midrule
1 Gbit/s, < 5 ms  &  1.71 & 1.54 & 70.0 & 253.6\\
100 Mbit/s, < 5 ms  &  1.66 & 1.49 & 56.4 & 182.0\\
100 Mbit/s, 100 ms &  1.23 & 1.11 & 19.7 & 112.2\\
\midrule
\multicolumn{5}{c}{\textsc{Petals} on 12 virtual servers}\\
\midrule
1 Gbit/s, < 5 ms  &  1.24 & 1.06 & 37.9 & 180.0\\
100 Mbit/s, < 5 ms  &   1.24 & 1.05 & 25.6 & 66.6\\
100 Mbit/s, 100 ms &   0.57 & 0.53 & 5.8  & 44.3\\
\midrule
\multicolumn{5}{c}{\textsc{Petals} on 14 real servers in Europe and North America}\\
\midrule
Real world  &  0.83 & 0.79 & 32.6 & 179.4 \\
\midrule
\multicolumn{5}{c}{Offloading, max. speed on 1x A100}\\
\midrule
256 Gbit/s &  0.18 & 0.18 & 2.7 & 170.3\\
128 Gbit/s &  0.09 & 0.09 & 2.4 & 152.8\\
\midrule
\multicolumn{5}{c}{Offloading, max. speed on 3x A100}\\
\midrule
256 Gbit/s &  0.09 & 0.09 & 5.1 & 325.1\\
128 Gbit/s &  0.05 & 0.05 & 3.5 & 226.3\\
\bottomrule
\end{tabular}}
\vspace{-10pt}
\end{table}

\subsection{Benchmarks}

We evaluate the performance of \textsc{Petals} by running BLOOM-176B in emulated and real-world setups. Our first setup consists of 3 local servers, each running on an A100 80GB GPU. This is an optimistic scenario that requires the least amount of communication.
In the second setup, we simulate 12 weaker devices by partitioning each A100-80GB into several virtual servers (3 large and 1 small). 
We evaluate the above setups with three network configurations: 1~Gbit/s with < 5 ms latency, 100 Mbit/s with < 5 ms latency and 100 Mbit/s with 100 ms latency\footnote{We simulate network conditions with \url{https://github.com/magnific0/wondershaper}, which uses \texttt{tc qdisc}}. The client nodes have 8 CPU cores and no GPU.


Next, we benchmark BLOOM in a real-world distributed setting with 14 smaller servers holding 2$\times$~RTX~3060, 4$\times$2080Ti, 2$\times$3090, 2$\times$A4000, and 4$\times$A5000 GPUs. These are personal servers and servers from university labs, spread across Europe and North America and connected to the Internet at speeds of 100--1000 Mbit/s. Four of the servers operate from under firewalls\footnote{We use the Circuit Relay protocol from libp2p to traverse NATs and firewalls, see \url{https://docs.libp2p.io/concepts/circuit-relay/}}.

In Table~\ref{tbl:experiments}, we report the performance of single-batch inference and parallel forward passes. For inference, performance does not depend much on bandwidth or sequence length but degrades with higher latency.
Parallel forward passes with large batches (used for fine-tuning and parallel inference) are affected by both bandwidth and latency.

We also test the effect of having multiple clients. For 12 servers with 100 Mbit/s bandwidth and 100 ms latency, if 8 clients run inference concurrently, each of them gets $\approx20\%$ slowdown compared to the case when it runs inference alone.



Additionally, we compare \textsc{Petals} with parameter offloading to run large models with limited resources \citep{zerooffload,rajbhandari2021zero}. For the offloading benchmark we calculate the maximum inference and forward training throughput to receive an upper bound on offloading performance. We base our offloading numbers on the best possible hardware setup for offloading: CPU RAM offloading via PCIe 4.0 with 16 PCIe lanes per GPU and PCIe switches for pairs of GPUs.

We calculate the maximum throughput for offloading as follows. In 8-bit, the model uses 1 GB of memory per billion parameters while PCIe~4.0 with 16 lanes has a throughput of 256 Gbit/s (or 128 Gbit/s if two GPUs are behind a PCIe switch). As such, offloading 176B parameters takes 5.5 seconds for a regular setup and 11 seconds for a multi-GPU setup. We assume an offloading latency of zero for the upper bound estimation.

These results are also shown in Table~\ref{tbl:experiments}. We can see that offloading is about an order of magnitude slower for single-batch inference compared to \textsc{Petals}. For the fine-tuning forward pass, offloading is competitive if multiple GPUs are used and the networking for \textsc{Petals} is limited to 100 Mbit/s or has high latency. In other cases, \textsc{Petals} offers higher throughput than offloading for training.
\section{Discussion and future work}\label{sect:discussion}\label{sect:incentives}

\paragraph{Incentives for peers to contribute.}
In \textsc{Petals}, peers using the client are not required to run a server. This may lead to an imbalance between supply (peers who dedicate GPUs to serve model layers) and demand (peers using the servers to perform inference or fine-tuning for their own needs) in the network.
One way to encourage users to serve model layers is to introduce a system of \textit{incentives}: peers running servers would earn special \textit{points}, which can be spent on high-priority inference and fine-tuning or exchanged for other rewards.

\paragraph{Privacy.} An important limitation of our approach is that peers serving the first layers of the model can use their inputs to recover input tokens. Thus, people working with sensitive data should limit their clients to only use trusted servers or, alternatively, set up their own isolated \textsc{Petals} swarm.

This limitation may be addressed in future using secure multi-party computing~\citep{evans2018pragmatic} or privacy-preserving hardware~\citep{nvidia-privacy}.

\paragraph{Security.}
We assume that servers in our system are run by many independent parties. In practice, some of them may turn out to be faulty and return incorrect outputs instead of the actual results of forward and backward passes. This may happen due to a malicious intent to influence other people's outputs or, when rewards are introduced (as described above), to earn a reward for serving layers without actually performing the calculations.

A possible way to address these issues would be to use an economically motivated approach.
Some servers may vouch for the correctness of their outputs (e.g., in exchange for increased inference price) by depositing a certain number of points as a pledge. Then, for each request, they announce a cryptographic hash of the input and output tensors, so anyone having the inputs can check whether the outputs are correct.

If someone finds a mismatch confirmed by a trusted third party, they can claim the server's pledge as a reward. In practice, it may be a client who suspects that they received wrong outputs or a ``bounty hunter'' sending requests to different servers in the hope of catching errors.
While this approach still leaves a chance of receiving wrong outputs, it makes cheating costly and creates an incentive to quickly expose the malicious servers.






\paragraph{Making changes to the main model.}
As discussed in Section~\ref{sect:design_training}, distributed parameter-efficient fine-tuning makes it easy for users to apply the base model to new tasks.
In Section~\ref{sect:design_ecosystem}, we also described how these updates can be easily shared and reused by others.
This capability provides a meaningful step towards \textit{collaborative} improvement of machine learning models~\citep{raffel2021call}:
as more and more users train the base model, it will effectively become more capable over time.


Furthermore, we might expect the model parameters that perform best on a specific task to change over time.
Similarly to version control systems for code, it would be useful to track versions of fine-tuned model parameters as they change.
A system for rapidly testing the performance of a set of parameters on ``living benchmarks''~\citep{dynabench,gehrmann2022gemv2,eval-harness} would be valuable to ensure that subsequent versions improve the desired capabilities.

Apart from adaptation to new tasks, it would also be useful to eventually update the main model.
Ideally, such updates could be tracked in a principled way.
Users of \textsc{Petals} could specify the versions of the model they want to use, and servers could indicate which versions they support.
Introducing a newer version of the model then reduces to adding a new group of layers, which then naturally supersedes older parameters based on the approach from Section~\ref{sect:networking}.
Similarly, fine-tuned adapters could be annotated with tags denoting the model version they are applicable for.
Such fine-grained model versioning is currently uncommon but would be straightforward to add to \textsc{Petals}.

\section{Conclusion}

This paper introduces \textsc{Petals}, a system for efficient collaborative inference and fine-tuning of large language models. We offer a user-friendly generation interface and a flexible API to access models served over the Internet. We use 8-bit compression that reduces the resource requirements to run very large models. In addition, we develop algorithms for reliable routing and load balancing.

With the release of this system, we hope to broaden access to LLMs and pave the road to applications, studies or research questions that were previously not possible or simply too expensive.





\section*{Ethics Statement}
This work introduces a general-purpose algorithm for decentralized inference of large models, aiming to simplify access to the latest research in deep learning. Thus, we do not envision any direct negative impacts from our research aside from granting the broader public an ability to interact with LLMs trained on uncurated web-crawled data. However, all models we serve are already in open access and thus can be exposed via APIs or other means.

\section*{Acknowledgements}

The authors thank Zheng-Xin Yong, Ilya Dimov, Yozh, Teven Le Scao, Stas Bekman, and Haokun Liu for helpful discussions. We also thank Teven Le Scao for his help in designing Figure~\ref{fig:overview}. A part of the experiments was conducted on a personal server of Elena Voita.

\clearpage

\bibliography{anthology,custom}
\bibliographystyle{acl_natbib}

\end{document}